# Advancing Retrieval-Augmented Generation for Persian: Development of Language Models, Comprehensive Benchmarks, and Best Practices for Optimization

Sara Bourbour Hosseinbeigi[1], Sina Asghari[2], Mohammad Ali Seif Kashani[3], Mohammad Hossein Shalchian[4], Mohammad Amin Abbasi[5]

[1]Department of Industrial and Systems Engineering, Tarbiat Modares University, Tehran, Iran
s.bourbour@modares.ac.ir

[2]Department of Computer Science, Iran University of Science and Technology, Tehran, Iran
sina_asghari@mathdep.iust.ac.ir

[3]Department of Computer Engineering, Sharif University of Technology, Tehran, Iran
ma.seifkashani@ce.sharif.edu

[4]Department of Computer Engineering, Sharif University of Technology, Tehran, Iran
mo.shalchian@sharif.edu

[5]Department of Computer Engineering, Iran University of Science and Technology, Tehran, Iran
m_abbasi1378@comp.iust.ac.ir

## ABSTRACT

*This paper examines the specific obstacles of constructing Retrieval-Augmented Generation (RAG) systems in low-resource languages, with a focus on Persian's complicated morphology and versatile syntax. The research aims to improve retrieval and generation accuracy by introducing Persian-specific models, namely MatinaRoberta (a masked language model) and MatinaSRoberta (a fine-tuned Sentence-BERT), along with a comprehensive benchmarking framework. Three datasets—general knowledge (PQuad), scientifically specialized texts, and organizational reports—were used to assess these models after they were trained on a varied corpus of 73.11 billion Persian tokens. The methodology involved extensive pretraining, fine-tuning with tailored loss functions, and systematic evaluations using both traditional metrics and the Retrieval-Augmented Generation Assessment (RAGAS) framework. The results show that MatinaSRoberta outperformed previous embeddings, achieving superior contextual relevance and retrieval accuracy across datasets. Temperature tweaking, chunk size modifications, and document summary indexing were explored to enhance RAG setups. Larger models like Llama-3.1 (70B) consistently demonstrated the highest generation accuracy, while smaller models faced challenges with domain-specific and formal contexts. The findings underscore the potential for developing RAG systems in Persian through customized embeddings and retrieval-generation settings and highlight the enhancement of NLP applications such as search engines and legal document analysis in low-resource languages.*

## KEYWORDS

*Retrieval-Augmented Generation, Large Language Models, Benchmarking, Persian, Sentence Embeddings*

# 1. INTRODUCTION

In natural language processing (NLP)[1], retrieval-augmented generation (RAG)[2] systems enhance generative outputs by incorporating external knowledge bases, addressing the limitations of standalone large language models (LLMs)[3] which often suffer from outdated or insufficient knowledge. Recent advancements highlight the potential of integrating high-quality retrieval mechanisms with fine-tuned language models to reduce hallucinations[4] and improve factual consistency. However, applying these techniques to languages with limited resources, such as Persian, presents unique challenges. Persian's rich morphology, flexible syntax, and scarcity of annotated resources necessitate specialized approaches to both retrieval and language modeling.

Despite advancements in improving LLMs through retrieval-augmented frameworks, their application to low-resource languages like Persian is underdeveloped. Significant barriers include a lack of domain-specific pretraining corpora and custom models for sentence-level embeddings, often leading to reliance on general-purpose embeddings from high-resource languages, which are not optimal for retrieval accuracy and generative coherence. Furthermore, there are limitations in the applicability of these frameworks due to the absence of reliable benchmarks for evaluating RAG systems for Persian in various domains. While there has been progress in developing Persian-specific LLMs, such as PersianLLama[5], these efforts primarily focus on language modeling and generation rather than integrating retrieval-augmented capabilities, leaving a significant gap in RAG system development for Persian.

This project addresses these challenges by developing Persian-specific models, including a Sentence-BERT[6] and a masked language model, aimed at optimizing retrieval processes in RAG systems. A robust framework is established to systematically assess their performance across formal, technical, and general knowledge settings. The study also examines how various retrieval and generation configurations—such as document chunking, temperature settings, and summary-based indexing—impact key performance metrics like generative faithfulness, contextual relevance, and retrieval accuracy.

By achieving these objectives, this research advances LLMs in low-resource contexts, providing insights transferable to other underrepresented languages. It focuses on three representative datasets for evaluating RAG systems in Persian: (1) the PQuad dataset[7] for general knowledge tasks, (2) a scientific-specialized dataset for technical content, and (3) formal organizational reports, each reflecting diverse linguistic challenges and contextual needs. The study evaluates the performance of Persian-specific masked language models and Sentence-BERT models alongside other state-of-the-art LLMs.

This research makes the following contributions:
1. Development of Persian-specific masked language and Sentence-BERT models, filling crucial gaps in NLP resources.
2. Establishment of a comprehensive benchmark for systematic evaluation of RAG systems in low-resource languages.
3. Identification of best practices for RAG optimization, offering practical insights into model configurations and trade-offs between retrieval and generation.

These contributions enhance the understanding of RAG systems, opening the door for improved applications in search engines, legal document analysis, and domain-specific retrieval systems by tailoring solutions to the linguistic features of Persian.

The paper is organized as follows: Section 2 reviews related work on retrieval-augmented generation systems and the application of LLMs to low-resource languages. Section 3 describes the development of MatinaRoberta, a Persian-specific masked language model, followed by Section 4, which introduces MatinaSentenceRoberta, a fine-tuned Sentence-BERT model for Persian retrieval tasks. Section 5 outlines the benchmarking methodology, datasets, and evaluation metrics used to assess RAG performance. Section 6 presents the results and discussion, analyzing model performance across diverse tasks and configurations. Finally, Section 7 concludes with a summary of contributions, implications for NLP in low-resource settings, and future research directions.

## 2 RELATED WORK

### 2.1 Searching for Best Practices in Retrieval-Augmented Generation

Xiaohua Wang[8] and colleagues explore the optimization of retrieval-augmented generation (RAG) systems through query-dependent retrieval processes. They analyze various components of the RAG pipeline, such as query classification, document retrieval, reranking, and summarization, in order to improve efficiency and accuracy. The authors emphasize the use of multimodal retrieval approaches, such as integrating text-to-image and image-to-text retrieval, particularly for question-and-answer tasks in visual domains. They propose best practices for balancing system performance and complexity by conducting systematic comparisons. This work is significant for the advancement RAG systems, because it provides insights that can be applied across multiple domains, including both multimodal and text-only contexts.

### 2.2. Benchmarking Large Language Models in Retrieval-Augmented Generation

Chen et al.[9] evaluate the performance of large language models in Retrieval-Augmented Generation tasks. The study performs a thorough analysis of key challenges that LLMs face while integrating external knowledge through retrieval. It establishes a Retrieval-Augmented Generation Benchmark (RGB) to assess four key capabilities required for effective RAG: noise robustness, negative rejection, information integration, and counterfactual robustness. The authors use this benchmark, to evaluate six state-of-the-art LLMs, revealing that while LLMs are somewhat robust to noise, they still struggle greatly with rejecting incorrect information, integrating multiple information sources, and handling counterfactual errors.

This work contributes to the ongoing endeavours improve RAG systems by identifying inefficiencies in current LLMs and providing a new benchmark for evaluation. It underscores the potential of retrieval to enhance LLM performance, particularly in mitigating hallucinations and addressing outdated knowledge. However, it also points out the challenges, particularly in terms of how LLMs handle external misinformation and integrate complex information from multiple documents. The findings of this work offer valuable insights for future research focused on improving the reliability and robustness of RAG systems.

### 2.3 CRUD-RAG: A Chinese Benchmark for RAG Systems

Lyu et al.[10] introduce a comprehensive benchmark for evaluating the performance of Retrieval-Augmented Generation systems, with a specific focus on Chinese language tasks. The authors divide RAG tasks into four CRUD operations: Create, Read, Update, and Delete, and design separate evaluation tasks for each. Examples include text continuation, question answering, multi-document summary, and hallucination modification. Their benchmark tackles the shortcomings of current RAG evaluations, which often focus solely on question-answering and overlook the broader range of applications. By taking into account numerous various components such as retrievers, chunk size, and embedding models, the research provides valuable insights for optimizing RAG systems across diverse tasks and application scenarios.

Unlike previous studies, this research provides a novel framework for RAG in low-resource languages, focusing on structural and morphological challenges and constraints given by Persian. By focusing on retrieval-generation pipelines, it sets a new benchmark for Persian NLP and has larger implications for RAG in other low-resource languages.

## 3 MATINAROBERTA

Effective Retrieval-Augmented Generation systems rely significantly on high-quality sentence embeddings to achieve precise retrieval and contextually relevant generation. Sentence-BERT plays a critical role in this pipeline, as it generates dense vector representations that capture semantic similarities between text segments, which is essential for accurate retrieval in RAG frameworks. However, the efficacy of Sentence-BERT is contingent upon the availability of a robust underlying language model that has been pretrained on a diverse and high-quality corpus of data.

To meet this need, we conducted continual pretraining of XLM-RoBERTa[11] Large on a vast corpus comprising 54.69 billion Persian tokens. The masked language modeling (MLM) for fine-tuning into a Sentence-BERT architecture. Notably, the Next Sentence Prediction (NSP) task was omitted, which is consistent with previous research that shows higher performance in models trained without NSP. This approach ensures the model captures Persian's intricate morphology and flexible syntax, offering a solid foundation for its further use in generating high-quality embeddings for RAG systems.

### 3.1 Corpus Details

The pretraining corpus was drawn from a variety set of Persian textual sources to ensure comprehensive coverage of linguistic and contextual variations, detailed in Table 1. Scientific articles contributed 3.55 billion tokens, providing formal and structured content. General, educational, and religious books added 2.84 billion tokens, reflecting a wide range of genres and domains. Informal and conversational language was caught by 2.34 billion tokens from social media, while Persian websites provided 14.78 billion tokens spanning news, blogs, and forums. Additionally, 49.6 billion tokens were taken from Common Crawl datasets, to ensure coverage of modern and dynamic language use. Together, these sources covered a wide range of Persian text types, enabling the model to generalize effectively. Table 1 below provides a full breakdown of the datasets and their corresponding token counts, illustrating the diverse textual sources used in the pretraining process.

Table 1. Token Distribution Across Pretraining Datasets.

| Dataset | # Tokens |
| --- | --- |
| Scientific articles | 3.55B |
| Books | 2.84B |
| Social Media | 2.34B |
| Persian Websites | 14.78B |
| Common Crawl | 49.6B |
| **Sum** | **73.11B** |

## 3.2 Preprocessing

To achieve good quality, the data was extensively pre-processed prior to training. This includes deduplication and noise removal, enhancing the corpus's overall integrity. These steps were critical to ensure the relevance and accuracy of the pretraining data, particularly for a low-resource language like Persian.

## 3.3 Training Procedure

The training process was built up to maximize efficiency while processing the extensive dataset. The MLM objective required the model to predict randomly masked tokens based on their surrounding context, which effectively captured nuanced word associations. This was particularly valuable for Persian, where the complex morphology and syntax provide distinct issues.

Over the course of one week, training was carried out on eight NVIDIA A800 GPUs, with DeepSpeed's Stage 0[12] optimization framework used to boost computational performance. The training process employed a maximum sequence length of 512 tokens and FP16 mixed precision for efficient computation. To optimize convergence the learning rate was adjusted to 5e-5, with a linear warm-up schedule. The batch size for training was configured at 30 per device, with a gradient accumulation of two steps, resulting in an effective batch size of 480. The evaluation batch size was set to eight per device.

The AdamW optimizer was used during the optimization process, using betas of (0.9, 0.999), and an epsilon of 1e-8. The learning rate was managed with a linear scheduler. The model was trained for one epoch, allowing comprehensive exposure to the entire dataset.

## 3.4 Applications and Impact

The MatinaRoberta model excels in a range of Persian NLP tasks, including text classification, question answering, semantic search, and contextual embedding generation. It is particularly effective in Retrieval-Augmented Generation systems, where the embeddings enable precise retrieval and factually grounded generation. This enhances applications like Persian-language search engines, knowledge assistants, and domain-specific retrieval, ensuring cutting-edge performance for low-resource languages. The results of its evaluation compared with other models can be found in Table 2.

Table 2. Results of Masked Language Models Evaluation.

| Model | Arman Emo | Pars-ABSA | Taghche | Snapp Food | Digikala | Digimag | Persian News | PQUAD | Reading Comprehension | DeepSentPars | PEYMA |
|---|---|---|---|---|---|---|---|---|---|---|---|
| MatinaRoberta | **56.54** | **74.92** | **59.34** | **88.7** | **72.59** | **96.37** | **98.05** | 86.82 | **56.82** | **83.21** | 85.65 |
| TookaBERT[13] | 52.87 | 74.65 | 58.17 | 87.26 | 66.56 | 93.91 | 96.95 | 86.73 | 44.89 | 82.96 | 86.09 |
| AriaBERT[14] | 38.23 | 74.59 | 58.58 | 87.93 | 67.06 | 92.38 | 97.63 | 83.14 | 37.98 | 73.21 | 35.78 |
| XLM-RoBERTa | 32.48 | 74.18 | 54.59 | 86.28 | 63.06 | 91.58 | 96.9 | **87.6** | 42.55 | 73.39 | **87.94** |
| mBERT | 6.74 | 68.15 | 54.31 | 86.18 | 61.06 | 69.69 | 90.17 | 85.94 | 49.63 | 55.78 | 65.32 |

# 4 MATINASROBERTA

Following the masked language modeling pretraining, the model was fine-tuned to enhance its performance within RAG systems, namely as a similarity-based text embedding model. This fine-tuning process focused on aligning embeddings for semantically related text pairs while effectively distinguishing dissimilar pairs, which is crucial for improving retrieval accuracy and generation quality in RAG pipelines. The process utilized a diverse set of datasets that were carefully curated to cover a wide range of linguistic and semantic tasks, including question-answer matching, entailment classification, paraphrase detection, and triplet-based semantic similarity. These efforts significantly optimized the model's ability to retrieve and integrate relevant information, enabling the generation of contextually accurate and semantically rich responses, establishing it as a state-of-the-art solution within RAG frameworks.

Key datasets incorporated into the fine-tuning included multilingual resources such as the Miracle Project's triplet datasets, covering English, Arabic, and Persian subsets. Additionally, the ParsiNLU[15] datasets, such as question pair similarity and entailment classification tasks, added valuable linguistic and contextual changes. Domain-specific Persian datasets played a critical role in enriching the model's comprehension of specialized settings. These included QA pairs from the PQuad collection, paraphrase pairs, Wikipedia-derived triplets and paired data, as well as datasets focusing on religious, legal, educational, and tourism-related texts. Instructional Persian datasets and bilingual Persian-English translation pairing improved the model's ability to handle a variety of language constructions and semantic relationships.

The fine-tuning procedure employed various loss functions tailored to the datasets' structures and tasks. For datasets having anchor-positive pairs, Multiple Negatives Ranking Loss was used to leverage other samples within a batch as negatives, optimizing for semantic similarity learning. Contrastive Loss was applied to datasets that required binary classification of semantic similarity, whereas Softmax Loss was used for tasks involving three-class entailment classification. For datasets containing anchor, positive, and negative samples, Triplet Loss was utilized to ensure that embeddings for anchors and positives were closer than those for anchors and negatives. With these loss functions, the model was able to effectively adapt to a wide range of semantic tasks.

Embeddings generated during this process were dense 1024-dimensional vectors, obtained by averaging the token embeddings from the transformer layers. The architecture and training approach were inspired by Sentence-BERT, which resulted in efficient and high-quality semantic representations. Training was carried out over 7 days using 8 NVIDIA A800 GPUs, with DeepSpeed employed to optimize computational resources. Specific training parameters, detailed in Table 3, provided a structured and efficient training regimen.

Table 3. Token Distribution Across Pretraining Datasets.

| Config | Value |
| --- | --- |
| Train batch_size | 30 |
| Gradient accumulation steps | 2 |
| Weight decay | 0.01 |
| Num train epochs | 4 |
| Lr scheduler type | polynomial |
| Warmup ratio | 0.4 |
| FP16 | True |

# 5 BENCHMARKING RAG

This section outlines the experimental design and methods employed to evaluate the performance of Retrieval-Augmented Generation in Persian. The study systematically tested various embedding models and large language models across three datasets—general knowledge, scientific-specialized texts, and formal organizational documents—aiming to identify the most effective strategies for optimizing RAG systems in the context of Persian language. This evaluation addresses not only linguistic challenges but also the adaptability of models to different content types.

## 5.1 Evaluation Datasets

To thoroughly evaluate the retrieval and generation capabilities of the models, three distinct datasets were employed, each reflecting a different type of textual content and linguistic complexity.

### 5.1.1 General Knowledge Dataset

The general knowledge dataset was drawn from PQuad, a Persian-language reading comprehension dataset sourced from Wikipedia articles. This dataset contains approximately 80,000 questions, 25% of which are classified as unanswerable. It is divided into training (63,994), validation (7,976), and test (8,002) subsets. The wide range of topics covered by Wikipedia makes this dataset particularly suitable for evaluating the generalization ability of RAG models in retrieving and generating information from diverse, open-domain content. The variety of topics helps test the robustness of models in handling Persian text with various contextual demands and linguistic features.

### 5.1.2 Scientific-Specialized Dataset

The scientific-specialized dataset was created using content from the Persian-language textbook *General Physical Education[16]*. This textbook contains comprehensive information on the philosophy of physical education, exercise physiology, and health sciences, making it an ideal source for evaluating models' performance in handling domain-specific language. To generate the dataset, the text was preprocessed by cleaning redundant sections and extracting meaningful multiple-choice questions (MCQs) using GPT-4 model. This dataset assesses how well LLMs can retrieve and process specialized terminology and knowledge, a critical aspect when applying RAG in technical or academic fields. The structured format of the text, including specialized terms and concepts, presented additional challenges for retrieval accuracy, making this dataset a rigorous test for the models.

### 5.1.3 Organizational Report Dataset

The third dataset was built using the Fundamental Transformation Document of Education in Iran, a formal policy document detailing key strategies and frameworks for overhauling the Iranian education system by the year 1404 (2025). This document contains structured, formal language, along with socio-political and cultural terminology, making it an important test case for evaluating how well models can retrieve and generate information from highly formalized and context-specific texts. MCQs generated from this document using GPT-4o assessed the models' ability to process lengthy, policy-oriented texts with intricate cultural and organizational references. The formal structure of this document, combined with its specialized vocabulary, created a challenging environment for RAG systems to navigate, testing both their understanding of formal syntax and their ability to contextualize information.

## 5.2 Embedding Model Evaluation

The embedding models evaluated in this study include MatinaSRoberta, LaBSE[17], L12-V2, Qwen2-7[18], and Alibaba/gte. These models were selected based on their proven performance in related tasks involving low-resource languages and their ability to generate embeddings that capture complex linguistic features in Persian. Additionally, the models chosen are compatible with the Sentence-Transformers framework, ensuring support for creating high-quality sentence embeddings. This compatibility enhances their effectiveness in handling retrieval and generation tasks in the RAG system. Each model was evaluated across the three datasets to determine its effectiveness in retrieving relevant content in Persian, with a focus on retrieval accuracy in a low-resource context where available data is often limited, and linguistic structures are complex.

## 5.3 Experimental Setup

### 5.3.1 Baseline Evaluation of Large Language Models

The capabilities of the LLMs were evaluated using a Multiple-Choice Question Answering (MCQA), built on the LlamaIndex framework, which was applied to each of the three datasets (mentioned above). The models evaluated include LLaMA 3.1 (8B & 70B)[19], Qwen 2 (7B & 72B), Gemma 1.1[20], and Gemma 2[21]. These models were selected based on their superior performance in other low-resource languages, suggesting their potential to handle the linguistic complexity of Persian, which includes flexible word order, morphological richness, and a diverse vocabulary. By testing these models under static conditions within the RAG framework, we established a baseline for their ability to generate accurate, contextually appropriate answers based on retrieved information. This baseline offers a foundation for understanding how well these models can generalize across multiple document types and language domains. In Table 4, we present the base parameters used for evaluations, providing a comprehensive overview of the configuration for each model during the baseline tests.

Table 4. The base parameters used for evaluations.

| Config | Value |
| --- | --- |
| max tokens | 2048 |
| chunk size | 1024 |
| chunk overlap | 256 |
| similarity top k | 5 |
| temperature | 0.25 |

### 5.3.2 Temperature Tuning

Temperature tuning was applied to refine the performance of the selected LLMs, with a focus on the LLaMA 3.1 8B model. The temperature parameter controls the degree of randomness in the model's outputs—lower temperatures generate more deterministic, precise answers, while higher temperatures encourage diversity in responses. In the context of Persian, where sentence structure can be more flexible than in other languages, temperature tuning plays a critical role in balancing precision and variability.

Four temperature settings (0, 0.25, 0.5, and 0.75) were tested to determine the optimal balance between accuracy and diversity in responses. Lower temperature settings (close to 0) provided more precise but less varied answers, which occasionally led to issues in handling Persian's flexible word order and polysemous words. Higher temperature, on the other hand, introduced more variability in responses but at the cost of factual consistency.

### 5.3.3 Chunk Size Testing

The effect of chunk size on retrieval performance was also examined, particularly in relation to the varying content structures of the three datasets (mentioned above). Using the LLaMA 3.1 8B model, chunk sizes of 512, 1024, and 2048 tokens were tested to assess how retrieval accuracy and model performance varied with changes in text segmentation.

Smaller chunk sizes (e.g., 512 tokens) generally resulted in more precise retrievals, as they allowed the model to focus on finer-grained sections of text. However, this precision came at the cost of higher computational demands, as smaller chunks required more iterations to process. Larger chunk sizes (e.g., 2048 tokens) improved computational efficiency but reduced retrieval precision, particularly in formal and technical texts where nuances in small sections could be critical.

### 5.3.4 Document Summary Indexing

To further optimize the retrieval process, we tested a Document Summary Indexing method. This approach involved using an LLM during the indexing phase to generate concise summaries of each document. These summaries, along with smaller text chunks (nodes), were stored and later retrieved during query processing. This method diverges from traditional retrieval approaches by reducing computational load and improving response time, particularly when handling large or complex documents.

When a query was issued, the model first evaluated document summaries to assess relevance before retrieving and analyzing the associated text chunks. This approach enabled faster and more efficient retrieval, especially for longer documents like *the Fundamental Transformation Document of Education in Iran*.

Additionally, a query classification prompt was implemented to distinguish between general queries (which required broader summaries) and specific queries (which demanded detailed retrieval). This distinction helped optimize retrieval accuracy and efficiency by ensuring that queries were matched with the most relevant sections of the document, thereby enhancing both precision and relevance in the retrieval process.

## 5.4 RAGAS Framework

In this study, we employed the Retrieval-Augmented Generation Assessment (RAGAS)[22] framework to evaluate the performance of various Retrieval-Augmented Generation models. RAGAS offers a structured, automated evaluation across multiple metrics, facilitating a comprehensive assessment of both retrieval and generation capabilities. This framework was selected for its adaptability in evaluating models without relying heavily on ground truth data, which is scarce for low-resource languages like Persian. Traditional benchmarks often do not exist for these languages, so a more flexible and automated assessment method like RAGAS was necessary.

While the MCQA framework was used to assess generative capabilities, RAGAS provided a more holistic evaluation of both retrieval and generation. Key metrics in RAGAS include factual consistency, context relevance, retrieval accuracy, and model precision in handling queries. This dual evaluation approach—utilizing both MCQA and RAGAS—ensured that a wide range of tasks was covered, from highly structured multiple-choice tests to broader generation and retrieval performance metrics. This allowed us to assess the models' effectiveness in real-world scenarios, where both factual accuracy and contextual relevance are critical.

## 5.5 Evaluation Metrics

### 5.5.1 Evaluation Metrics for Embedding Models

To gauge the performance of each embedding model, the following metrics were used:
- First Result Accuracy: The percentage of queries where the correct answer was retrieved as the first result.
- Second Result Accuracy: The percentage of queries where the correct answer appeared in the second result.
- Third Result Accuracy: The percentage of queries where the correct answer was found in the third result.
- Total Retrievals: The total number of correct retrievals across all results.
- Overall Score: A composite metric calculated based on the number of correct retrievals, offering a comprehensive view of each model's performance.

To calculate the Overall Score, we assigned different weightings to each result: First Result accuracy was given a factor of 3, Second Result accuracy was assigned a factor of 2, and Third Result accuracy was weighted with a factor of 1. These weighted scores offer a more nuanced evaluation of the models' retrieval capabilities, prioritizing higher-ranking results in the assessment of their effectiveness. These metrics were designed to offer insights into how well the embedding models retrieved pertinent information across various types of text, reflecting both the generalizability and specificity of the models.

### 5.5.2 Evaluation Metrics for Large Language Models

In this study, the evaluation of large language models was conducted using both the RAGAS framework and one additional metric for the Multiple-Choice Question Answering task. The metrics applied include:
- Accuracy: This metric evaluates the LLMs' performance on the Multiple-Choice Question Answering task. Accuracy is defined as the percentage of correct answers generated out of the total number of questions. This metric provides a direct measure of the LLMs' ability to comprehend and generate correct answers based on retrieved information, serving as a key performance indicator for the models in structured question-answering contexts.
- Faithfulness: This metric assesses the factual consistency of the generated answer with the retrieved context. It is calculated by comparing the generated answer with the retrieved context, with scores ranging from 0 to 1. Higher values indicate stronger alignment between the generated response and the context, reflecting greater factual accuracy.
- Answer Relevance: This metric evaluates how well the generated answer responds to the given prompt. A higher score is assigned to answers that are complete, concise, and free of redundancy. In contrast, a lower score reflects answers that are incomplete or contain irrelevant information. Answer Relevance is computed based on the question, retrieved context, and generated answer.

- Context Precision: Context Precision measures whether the retrieved context correctly prioritizes relevant items, ranking them higher in the result set. It assesses how effectively the retrieval process surfaces the most pertinent chunks of information. Scores range between 0 and 1, with higher scores indicating that relevant items are ranked more effectively.
- Context Recall: This metric assesses how well the retrieved context supports the ground truth answer, treated as the reference. It is calculated by determining the degree to which claims in the ground truth answer are attributable to the retrieved context. Scores range from 0 to 1, with higher values reflecting stronger performance. A reference-free version of this metric, context utilization, is also available for scenarios where ground-truth answers are not explicitly provided.

These metrics provide a comprehensive view of the LLMs' effectiveness in generating accurate, relevant answers and retrieving supportive context, ensuring both answer quality and factual consistency.

# 6 RESULTS AND DISCUSSION

This section presents the evaluation outcomes for both embedding models and large language models across three datasets: PQuad (general knowledge), scientific-specialized, and organizational reports. Each model's performance was evaluated through various metrics, including retrieval accuracy, context precision, and answer relevance, to determine its effectiveness in generating and retrieving Persian text. The following sections summarize key findings, with performance metrics detailed in the accompanying tables.

## 6.1 Embedding Models Evaluation

The embedding models were assessed on their ability to retrieve relevant information from the three datasets (mentioned earlier). The results of this evaluation, including retrieval accuracy and other performance metrics for each model, are presented in Table 5. Retrieval accuracy was measured for each model based on its performance in returning relevant results in the first, second, or third rank. This ranking system provided insights into the model's efficiency across different contexts, ranging from open-domain content to specialized technical texts.

MatinaSRoberta model consistently demonstrated superior performance across all three datasets, particularly in the PQuad and scientific-specialized datasets, where it outperformed models such as LaBSE and L12-V2. This performance can be attributed to MatinaSRoberta's ability to better handle Persian's linguistic features, including flexible sentence structures, morphological variations, and agglutination.

Table 5. Performance of Sentence-Transformer Models Across Datasets.

| Model | PQUAD | | | | Scientific-Specialized | | | | Organizational Report | | | |
|---|---|---|---|---|---|---|---|---|---|---|---|---|
| | 1th | 2th | 3th | Avg | 1th | 2th | 3th | Avg | 1th | 2th | 3th | Avg |
| MatinaSRoberta | 8231 | 475 | 142 | 47.70 | 766 | 206 | 22 | 42.23 | 356 | 150 | 32 | 21.02 |
| Ahd | 7231 | 604 | 181 | 42.70 | 715 | 206 | 76 | 40.70 | 412 | 106 | 52 | 22.52 |
| LaBSE | 4713 | 1041 | 457 | 30.8ه | 403 | 172 | 91 | 25.41 | 412 | 106 | 52 | 22.46 |
| L12-V2 | 4087 | 843 | 413 | 26.56 | 541 | 111 | 46 | 29.23 | 198 | 95 | 30 | 12.22 |
| Qwen2-7 | 3979 | 608 | 278 | 24.85 | 87 | 83 | 83 | 7.88 | 46 | 51 | 85 | 4.879 |
| Alibaba/gte-large | 1341 | 476 | 313 | 9.78 | 290 | 121 | 107 | 18.84 | 310 | 67 | 120 | 17.77 |
| Alibaba/gte | 814 | 359 | 256 | 6.32 | 298 | 43 | 58 | 16.04 | 174 | 66 | 46 | 10.51 |

MatinaSRoberta's success in PQuad dataset is largely due to its capacity for handling diverse, open-domain topics. Persian's flexible word order and varied morphology can complicate text retrieval, but MatinaSRoberta appeared well-suited to generalizing across these variations. In contrast, LaBSE and OpenAI embeddings, struggled with generalization, often failing to retrieve the most contextually appropriate content or misclassifying related content as irrelevant.

In the scientific-specialized dataset, which contains domain-specific jargon and technical terms, MatinaSRoberta outperformed other models by maintaining high precision in retrieving relevant information. Its success may be attributed to its stronger ability to process specialized terminology in Persian, effectively handling complex structures. In comparison, LaBSE model exhibited weaknesses in retrieving highly specific information, possibly due to their training data being skewed toward more general language use.

For the organizational report dataset, derived from formal documents with socio-political and cultural references, MatinaSRoberta once again outperformed LaBSE and OpenAI embeddings. This dataset posed significant challenges due to its formal language structure and context-specific terminology. MatinaSRoberta's ability to navigate these intricacies indicates that it is well-suited for retrieving from structured and formal texts, while LaBSE struggled, potentially missing important cultural nuances.

The MatinaSRoberta onsistently demonstrated the highest performance across most datasets, highlighting the importance of using embeddings tailored to the linguistic complexities of Persian. These findings underscore the value of specialized embeddings for optimizing retrieval in low-resource languages like Persian, particularly for domain-specific and formal contexts.

## 6.2 Evaluation of Large Language Models in RAG

### 6.2.1 Baseline Evaluation

The baseline evaluation assessed the generative capabilities of large language models in producing accurate and contextually relevant answers based on the information retrieved by embedding models. Table 6 summarizes the performance of various LLMs, highlighting the superior generative capabilities of larger models, such as LLaMA-3.1 (70B) and Qwen2 Instruct (72B) across all datasets.

In the PQuad dataset, which focuses on general knowledge, larger models like LLaMA-3.1 (70B) achieved notably higher accuracy in generating contextually appropriate responses. Their broader parameter size allowed these models to manage the diversity of topics more effectively, leading to precise answers based on the retrieved information. In contrast, smaller models like Qwen2 (7B) and Gemma 1.1 struggled to handle the wide range of content, showing limitations in managing the broad domain coverage required for general knowledge tasks.

Table 6. Baseline Evaluation of Large Language Models.

| Model | PQUAD | Scientific-Specialized | Organizational Report |
|---|---|---|---|
| Lamma-3.1-70B | 93.53 | 89.81 | 89.06 |
| Qwen2 Instruct - 72B | 91.98 | 78.88 | 85.27 |
| Gemma 1.1 it | 78.00 | 74.41 | 75.11 |
| Qwen2-7B | 76.30 | 60.62 | 82.73 |
| Lamma-3.1-8B | 74.89 | 76.92 | 84.55 |
| Lamma3 - 8B | 72.03 | 59.15 | 81.03 |

For the scientific-specialized dataset, LLaMA-3.1 (70B) maintained its advantage by generating accurate responses that aligned well with the technical language and domain-specific requirements of the dataset. The precision of these larger models in both retrieving and generating relevant answers demonstrated their ability to handle the complexity of Persian scientific texts. However, smaller models exhibited noticeable gaps in generating accurate responses, frequently misinterpreting specialized terminology or producing factually inconsistent answers.

The organizational report dataset, characterized by formal language and policy-oriented content, presented additional challenges. In this case, LLaMA-3.1 (70B) and Qwen2 Instruct (72B) once again outperformed smaller models by generating responses that accurately reflected the formal and structured nature of the documents. These larger models successfully incorporated the socio-political nuances present in the data, while the smaller models displayed clear limitations when handling such highly formalized text structures.

In conclusion, larger LLMs consistently outperformed smaller ones across all datasets. This evaluation highlights the importance of model size and complexity, particularly for tasks requiring high precision in domain-specific and formal contexts. For applications in legal, academic, or policy research, larger models like LLaMA-3.1 (70B) are better equipped to handle the intricacies of Persian-language content.

### 6.2.1 Temperature Testing

Temperature testing was conducted on the LLaMA-3.1 8B model to evaluate the impact of randomness on text generation (Table 7). The results indicated that a temperature setting of 0.25 provided the optimal balance between accuracy and response diversity across all datasets. As temperatures increased, randomness also increased, which negatively impacted the precision of the generated responses, particularly in more structured or formal datasets like organizational reports. Higher temperatures (e.g., 0.75) resulted in less precise and more diverse outputs, often sacrificing the factual consistency needed in formal and specialized contexts.

Table 7. Impact of Temperature on Model Performance.

| Temperature | PQUAD | Scientific-Specialized | Organizational Report |
|---|---|---|---|
| 0 | 73.25 | 80 | 83.87 |
| 0.25 | 72.83 | 82.25 | 84.87 |
| 0.5 | 72.53 | 76.92 | 84.55 |
| 0.75 | 72.10 | 71.66 | 81.91 |

### 6.2.2 Chunk Size Testing

The effect of chunk size on retrieval performance was also evaluated (Table 8), particularly in relation to the different content structures of the datasets. The chunk size testing revealed that a smaller chunk size of 512 tokens produced the best overall results, especially in formal and structured datasets like the organizational report dataset. Smaller chunks allowed for more precise retrieval of relevant information, maintaining a higher degree of contextual relevance. In contrast, larger chunk sizes (e.g., 2048 tokens) resulted in decreased precision, particularly in specialized datasets, where granular information was critical for maintaining retrieval accuracy.

Table 8. Impact of Chunk Size on Model Performance.

| Chunk Size | PQUAD | Scientific-Specialized | Organizational Report |
|---|---|---|---|
| 512 | 72.13 | 75.47 | 92.62 |
| 1024 | 72.41 | 78.18 | 84.31 |
| 2048 | 70.37 | 65.82 | 84.38 |

### 6.2.3 Document Summary Index Testing

The use of a document summary index was also tested (Table 9) to optimize the retrieval process. The results demonstrated that document summaries significantly improved retrieval accuracy, particularly in general and formal datasets like PQuad and the organizational report dataset. The document summary indexing approach reduced the computational load and enhanced the model's ability to retrieve relevant sections of the text more efficiently. The combination of document summaries and query classification led to improved retrieval efficiency and accuracy for complex queries that required information from multiple sections of the document.

Table 9. Impact of Document Summary Indexing on Model Performance.

| With/Without Summary | PQUAD | Scientific-Specialized | Organizational Report |
|---|---|---|---|
| With Summary | 77.47 | 65.11 | 87.31 |
| Without Summary | 71.87 | 64.47 | 84.07 |

## 6.3 RAGAS Results

The RAGAS framework provided a comprehensive evaluation of both retrieval and generation performance across all datasets. The results revealed significant variation in model performance, with larger models like LLaMA-3.1 (70B) consistently demonstrating the highest performance in terms of answer relevancy and context recall across all datasets.

As shown in Table 10 for the PQuad dataset, LLaMA-3.1 (70B) achieved the highest answer relevancy (0.7455) and context recall (0.9205), making it the best-performing model for open-domain tasks. In contrast, smaller models like Gemma 1.1 struggled with generating coherent and relevant responses, despite having strong retrieval capabilities.

In the Scientific-Specialized RAGAS results, detailed in Table 11, LLaMA-3.1 (70B) continued to lead, demonstrating the highest faithfulness (0.7322) and context precision (0.8138). Interestingly, LLaMA-3.1(8B) achieved the highest context precision (0.9435), although its lower faithfulness suggests that, while it excelled at retrieving relevant content, it faced challenges in generating factually consistent responses. The Gemma models continued to struggle in this dataset, particularly with answer relevancy, confirming their difficulty in handling complex and technical language.

For the Organizational Report dataset, as outlined in Table 12, LLaMA-3.1 (70B) excelled once again, demonstrating strong answer relevancy (0.7507) and faithfulness (0.7388). In contrast, the Gemma models, performed poorly in this dataset, especially in answer relevancy (0.2919), despite maintaining decent context recall. This result highlights the Gemma models' ongoing difficulty in generating factually grounded and relevant answers for formal and highly structured documents.

In summary, the RAGAS results, clearly indicate that LLaMA-3.1 (70B) consistently outperformed other models across all datasets, demonstrating strong retrieval capabilities combined with high faithfulness and answer relevancy. The Qwen2 models followed closely, performing well in both general and formal contexts, but with slightly weaker performance in specialized content. While the Gemma models performed well in retrieval, they faced significant challenges in generating relevant and factually accurate responses, especially in more complex and formal datasets. These results emphasize the importance of further optimization for smaller models, particularly when applied to formal and domain-specific tasks in low-resource languages like Persian.

Table 10. PQUAD RAGAS.

| Model | Context Precision | Faithfulness | Answer Relevancy | Context Recall |
|---|---|---|---|---|
| Lamma-3.1-70B | 0.7750 | 0.6357 | 0.7455 | 0.9205 |
| Qwen2-Instruct-72B | 0.7750 | 0.6125 | 0.7026 | 0.9015 |
| Gemma 1.1 it | 0.7750 | 0.5410 | 0.3779 | 0.9281 |
| Gemma 2 it | 0.7750 | 0.5524 | 0.4234 | 0.9146 |
| Qwen2-7B | 0.7750 | 0.5500 | 0.6535 | 0.9021 |
| Lamma-3.1-8B | 0.7750 | 0.6412 | 0.7233 | 0.9385 |
| Lamma3-8B | 0.7750 | 0.6313 | 0.6938 | 0.9177 |

Table 11. Scientific-Specialized RAGAS.

| Model | Context Precision | Faithfulness | Answer Relevancy | Context Recall |
|---|---|---|---|---|
| Lamma-3.1-70B | 0.8138 | 0.7322 | 0.6317 | 0.8314 |
| Qwen2-Instruct-72B | 0.8138 | 0.6928 | 0.5728 | 0.8309 |
| Gemma 1.1 it | 0.8138 | 0.6035 | 0.4656 | 0.8429 |
| Gemma 2 it | 0.8138 | 0.6142 | 0.4452 | 0.8223 |
| Qwen2-7B | 0.8138 | 0.6101 | 0.4835 | 0.8380 |
| Lamma-3.1-8B | 0.8138 | 0.6412 | 0.7233 | 0.9385 |
| Lamma3-8B | 0.8138 | 0.6313 | 0.5151 | 0.8443 |

Table 12. Organization Report RAGAS.

| Model | Context Precision | Faithfulness | Answer Relevancy | Context Recall |
|---|---|---|---|---|
| Lamma-3.1-70B | 0.5355 | 0.7388 | 0.7507 | 0.8228 |
| Qwen2-Instruct-72B | 0.5355 | 0.7081 | 0.7214 | 0.8014 |
| Gemma 1.1 it | 0.5305 | 0.6173 | 0.3819 | 0.8291 |
| Gemma 2 it | 0.5305 | 0.6371 | 0.2919 | 0.8145 |
| Qwen2-7B | 0.5355 | 0.6942 | 0.6166 | 0.8080 |
| Lamma-3.1-8B | 0.5305 | 0.6926 | 0.7103 | 0.8308 |
| Lamma3-8B | 0.5355 | 0.5548 | 0.6085 | 0.8156 |

# 7   Conclusion

This study addressed the significant challenges associated with extending Retrieval-Augmented Generation (RAG) frameworks to low-resource languages, with a specific focus on Persian. Recognizing the linguistic complexities and the scarcity of annotated datasets, this research aimed to develop Persian-specific language models, establish comprehensive benchmarks, and identify best practices for optimizing RAG systems. These contributions hold critical value in advancing NLP for underrepresented languages and expanding the applicability of retrieval-enhanced frameworks.

The research presented the development of MatinaRoberta and MatinaSRoberta models, designed to capture the nuances of Persian's rich morphology and syntax. These models demonstrated superior performance in retrieval tasks compared to existing state-of-the-art embeddings. The introduction of a systematic benchmark provided a robust means of evaluating RAG systems across general, technical, and formal domains, further highlighting the models' ability to bridge gaps in existing NLP resources. Additionally, the results underscored the importance of tailoring model architectures and retrieval-generation pipelines to meet the demands of low-resource languages, offering insights transferable to other underrepresented linguistic contexts.

The implications of these findings are manifold. Theoretically, they contribute to the growing body of knowledge on enhancing RAG systems through language-specific adaptations. Practically, they lay the groundwork for improving domain-specific applications, such as search engines, legal document analysis, and educational content retrieval in Persian. Moreover, the insights derived from model evaluations and configurations—such as temperature tuning, chunk size optimization, and the use of document summary indexing—provide actionable guidelines for practitioners and policymakers seeking to enhance RAG systems in resource-constrained environments.

Building on these findings, future research could focus on expanding the dataset coverage to include more diverse dialects and specialized content areas. Additionally, exploring multilingual and multimodal approaches could further enhance retrieval accuracy and generative quality. Investigating the scalability of the proposed methods to other low-resource languages would also be a valuable avenue for exploration, potentially contributing to broader advancements in global NLP.

In conclusion, this research marks a significant step forward in addressing the linguistic and computational challenges of applying RAG systems to low-resource languages like Persian. By combining innovative model development with rigorous evaluation frameworks, the study not only enhances the understanding of RAG in such contexts but also provides a foundation for ongoing innovation and development in the field.

# REFERENCES

[1] K. Chowdhary and K. Chowdhary, "Natural language processing," *Fundamentals of artificial intelligence,* pp. 603-649, 2020.
[2] Y. Gao *et al.*, "Retrieval-augmented generation for large language models: A survey," *arXiv preprint arXiv:2312.10997,* 2023.
[3] S. Minaee *et al.*, "Large language models: A survey," *arXiv preprint arXiv:2402.06196,* 2024.
[4] S. Tonmoy *et al.*, "A comprehensive survey of hallucination mitigation techniques in large language models," *arXiv preprint arXiv:2401.01313,* 2024.
[5] M. A. Abbasi, A. Ghafouri, M. Firouzmandi, H. Naderi, and B. M. Bidgoli, "Persianllama: Towards building first persian large language model," *arXiv preprint arXiv:2312.15713,* 2023.
[6] N. Reimers, "Sentence-BERT: Sentence Embeddings using Siamese BERT-Networks," *arXiv preprint arXiv:1908.10084,* 2019.
[7] K. Darvishi, N. Shahbodaghkhan, Z. Abbasiantaeb, and S. Momtazi, "PQuAD: A Persian question answering dataset," *Computer Speech & Language,* vol. 80, p. 101486, 2023.
[8] X. Wang *et al.*, "Searching for best practices in retrieval-augmented generation," *arXiv preprint arXiv:2407.01219,* 2024.
[9] J. Chen, H. Lin, X. Han, and L. Sun, "Benchmarking large language models in retrieval-augmented generation," in *Proceedings of the AAAI Conference on Artificial Intelligence*, 2024, vol. 38, no. 16, pp. 17754-17762.
[10] Y. Lyu *et al.*, "Crud-rag: A comprehensive chinese benchmark for retrieval-augmented generation of large language models," *arXiv preprint arXiv:2401.17043,* 2024.
[11] A. Conneau, "Unsupervised cross-lingual representation learning at scale," *arXiv preprint arXiv:1911.02116,* 2019.
[12] J. Rasley, S. Rajbhandari, O. Ruwase, and Y. He, "Deepspeed: System optimizations enable training deep learning models with over 100 billion parameters," in *Proceedings of the 26th ACM SIGKDD International Conference on Knowledge Discovery & Data Mining*, 2020, pp. 3505-3506.
[13] M. SadraeiJavaheri *et al.*, "TookaBERT: A Step Forward for Persian NLU," *arXiv preprint arXiv:2407.16382,* 2024.
[14] A. Ghafouri, M. A. Abbasi, and H. Naderi, "AriaBERT: A Pre-trained Persian BERT Model for Natural Language Understanding," 2023.
[15] D. Khashabi *et al.*, "Parsinlu: a suite of language understanding challenges for persian," *Transactions of the Association for Computational Linguistics,* vol. 9, pp. 1147-1162, 2021.
[16] M. A. Mohammad Salar Mazyar, Mohammad Mahdi Azaryan, *General Physical Education*. University of Jiroft, 2018.
[17] F. Feng, Y. Yang, D. Cer, N. Arivazhagan, and W. Wang, "Language-agnostic BERT Sentence Embedding," in *Proceedings of the 60th Annual Meeting of the Association for Computational Linguistics (Volume 1: Long Papers)*, 2022, pp. 878-891.
[18] A. Yang *et al.*, "Qwen2 technical report," *arXiv preprint arXiv:2407.10671,* 2024.
[19] A. Dubey *et al.*, "The Llama 3 Herd of Models," *arXiv preprint arXiv:2407.21783,* 2024.
[20] G. Team *et al.*, "Gemma: Open models based on gemini research and technology," *arXiv preprint arXiv:2403.08295,* 2024.
[21] G. Team *et al.*, "Gemma 2: Improving open language models at a practical size," *arXiv preprint arXiv:2408.00118,* 2024.
[22] S. Es, J. James, L. E. Anke, and S. Schockaert, "RAGAs: Automated Evaluation of Retrieval Augmented Generation," in *Proceedings of the 18th Conference of the European Chapter of the Association for Computational Linguistics: System Demonstrations*, 2024, pp. 150-158.